\documentclass[letterpaper, 10 pt, conference]{ieeeconf}  

\IEEEoverridecommandlockouts                              
\usepackage{graphics} 
\usepackage{epsfig} 
\usepackage{times} 
\usepackage{amsmath} 
\usepackage{amssymb}  
\usepackage{graphicx}
\usepackage{subfig}
\usepackage{float}
\usepackage{orcidlink}
\DeclareMathOperator{\clamp}{clamp}
\DeclareMathOperator{\round}{round}

\title{\LARGE \bf
A Hetero-Associative Sequential Memory Model Utilizing Neuromorphic Signals: Validated on a Mobile Manipulator}

\author{Runcong Wang \orcidlink{0009-0006-1019-9481}, Fengyi Wang \orcidlink{0000-0003-0306-8601} and Gordon Cheng \orcidlink{0000-0003-0770-8717}
\thanks{Runcong Wang, Fengyi Wang, and Gordon Cheng are with the Institute for Cognitive Systems, Technical University of Munich, 80333 Munich, Germany {\tt\small runcong.wang@tum.de; fengyi.wang@tum.de; gordon.cheng@ieee.org}}
}

\begin{document}

\maketitle
\thispagestyle{empty}
\pagestyle{empty}


\begin{abstract}
This paper presents a hetero-associative sequential memory system for mobile manipulators that learns compact, neuromorphic bindings between robot joint states and tactile observations to produce step-wise action decisions with low compute and memory cost. The method encodes joint angles via population place coding and converts skin-measured forces into spike-rate features using an Izhikevich neuron model; both signals are transformed into bipolar binary vectors and bound element-wise to create associations stored in a large-capacity sequential memory. To improve separability in binary space and inject geometry from touch, we introduce 3D rotary positional embeddings that rotate subspaces as a function of sensed force direction, enabling fuzzy retrieval through a softmax weighted recall over temporally shifted action patterns. On a Toyota Human Support Robot covered by robot skin, the hetero-associative sequential memory system realizes a pseudo-compliance controller that moves the link under touch in the direction and with speed correlating to the amplitude of applied force, and it retrieves multi-joint grasp sequences by continuing tactile input. The system sets up quickly, trains from synchronized streams of states and observations, and exhibits a degree of generalization while remaining economical. Results demonstrate single-joint and full-arm behaviors executed via associative recall, and suggest extensions to imitation learning, motion planning, and multi-modal integration.

\end{abstract}

\section{INTRODUCTION}
Recent advances in robot motion planning leverage cutting-edge AI models, particularly transformers, diffusion models, and neural memory-based architectures. These new approaches enable handling complex environments, trajectory generalizations, and multi-modal perceptions, pushing robotics automation into a higher level.

Transformer model \cite{vaswaniAttentionAllYou2023} excels in sequence modeling and attention mechanisms, making it suitable for processing trajectories as sequences of states or actions. A combination of multiple transformer models presents their powerful adaptability on a variety of tasks in zero-shot \cite{liFrameworkRoboticManipulation2025}. Diffusion model \cite{hoDenoisingDiffusionProbabilistic2020} demonstrates its dexterity and robustness in robot manipulation tasks \cite{chiVisuomotorPolicyLearning}. It leverages the framework of denoising diffusion processes, where the policy learns to iteratively refine noisy action sequences into coherent, demonstration-aligned behaviors based on observed states or goals \cite{yuanUnpackingIndividualComponents2024,carvalhoMotionPlanningDiffusion2024, seoPRESTOFastMotion2025}. By connecting the transformer model and diffusion model, we can exhibit advantages of their own model \cite{wangMTDPModulatedTransformer2025}. 

However, these benefits come with significant limitations. He et al. \cite{heDemystifyingDiffusionPolicies2025} pointed out that when a high-capacity model, such as the diffusion model \cite{hoDenoisingDiffusionProbabilistic2020}, is trained on a small dataset, it tends to memorize fixed patterns in the high-dimensional latent space and essentially degrades into a lookup table. A true generalization can only be achieved by training on a large dataset, but it is already beyond the current state of the art \cite{heDemystifyingDiffusionPolicies2025}. In addition, the extensive floating-point matrix multiplications involved in these models impose high demands on computational resources.. 

These challenges motivate the exploration of alternative paradigms for efficient representation and recall. Associative memory models, such as the Hopfield network, provide a complementary approach by directly storing and retrieving patterns. The dense associative memory \cite{krotovDenseAssociativeMemory2016, demircigilModelAssociativeMemory2017} pushes the capacity limits of the original Hopfield network \cite{hopfieldNeuralNetworksPhysical1982}, extending the storage capacity to an exponential level. Ramsauer et al. \cite{ramsauerHopfieldNetworksAll2021} generalize the dense associative memory to an equivalent transformer form, i.e, the Hopfield layer, showing that the Hopfield network can have the same ability as transformer models. Their application \cite{widrichModernHopfieldNetworks2020} validated that the modern Hopfield networks have excellent performance on large memory and classification of similar patterns.

The Hopfield network family continues to explore auto-associative memory, the ability to recall a complete pattern from a partial or corrupted cue. Hetero-associative memory \cite{chartierBidirectionalHeteroassociativeMemory2006, moralesEntropicHeteroAssociativeMemory2024}, by contrast, stores associations between different kinds of patterns. This broadens the scope from mere completion to cross-modal retrieval (e.g., state-observation), cue-response mapping, supervised key-value lookup, and translation between representational domains, enabling content-addressable recall even when the cue and target live in different feature spaces.
Chaudhry et al. \cite{chaudhryLongSequenceHopfield2023} combined the original Hopfield Network and Dense Associative Memory, introduced interaction vertex \cite{krotovDenseAssociativeMemory2016, demircigilModelAssociativeMemory2017} to the original Hopfield Network, and proposed an asymmetric Hopfield Network with large storage capacity named DenseNet. It shows its ability to store sequences of patterns with exponentially large storage capacity. However, with increased interaction vertices, the implicit winner-takes-all feature of the DenseNet shows its shortcomings in generalization. The sequence patterns can only be recalled exactly as they are. 

More recently, neuromorphic computing has become popular in recent years \cite{kudithipudiNeuromorphicComputingScale2025, schumanOpportunitiesNeuromorphicComputing2022, yaoSpikebasedDynamicComputing2024}. Neuromorphic computing draws inspiration from biological nervous systems that transmit information via binary spikes. By encoding information in binary form, the neuromorphic architecture is computationally more efficient and consumes less memory and computational resources \cite{christensen20222022}. 

\begin{figure*}[!t] 
    \centering
        \includegraphics[width=0.95\textwidth]{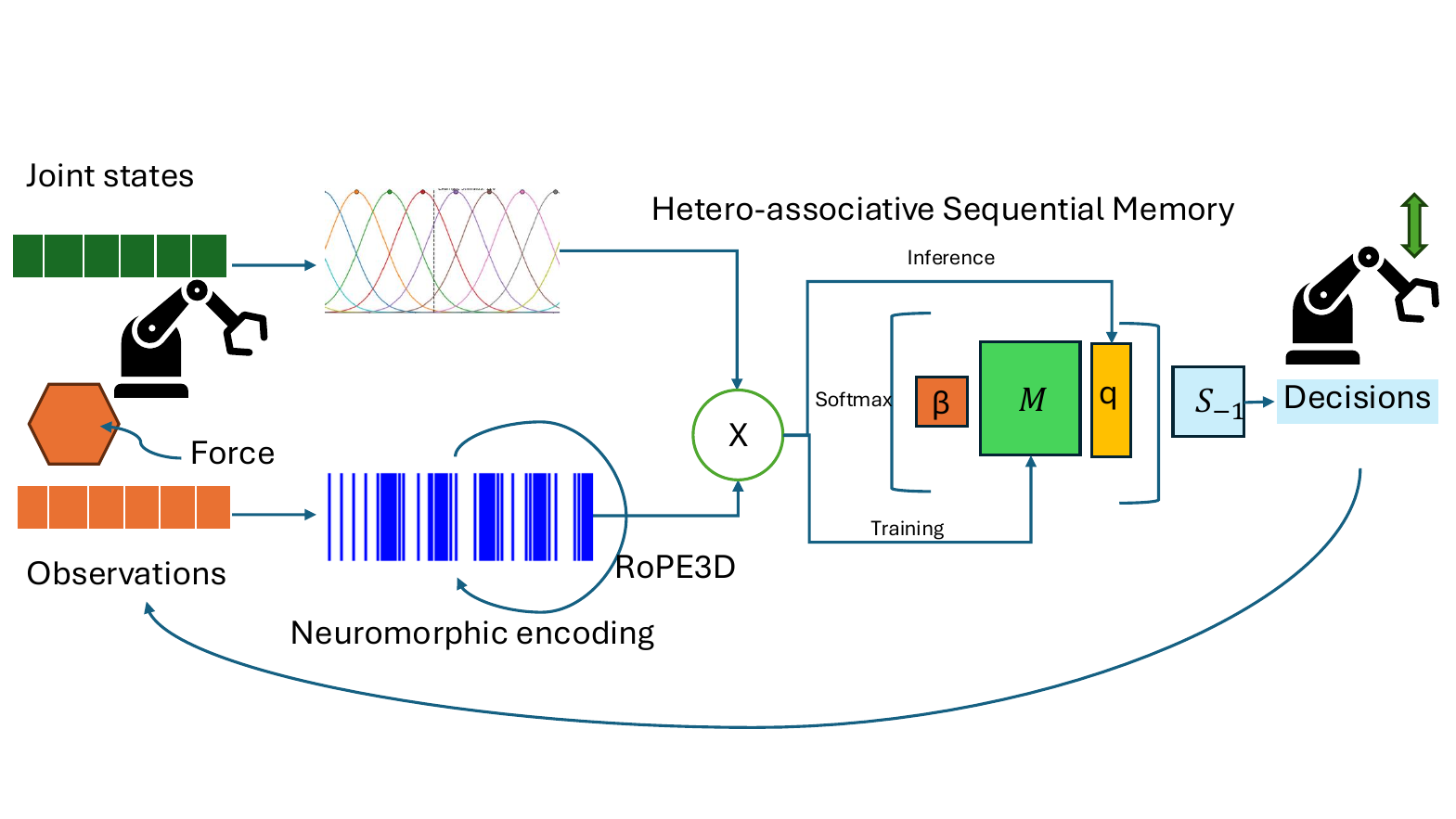} 
        \caption{Overview of the structure of hetero-associative sequential memory. The joint states and the observations are encoded and embedded into binary vectors in a high-dimensional space. For the training process, the memory matrix K stores associations between joint states and observations. }
        \label{fig:structure}
\end{figure*}

However, the binary vector space presents a significant challenge to the hetero-associative sequential memory, which conducts association search through the computation of inner products between vector pairs. The multiplicative interaction with zero elements causes the information annihilation, presenting a fundamental limitation for association search. In addition, the strong positional sensitivity of the binary data requires a position-dependent encoding method. To overcome these challenges, we introduce a 3D variety of Rotary Positional Embedding \cite{suRoFormerEnhancedTransformer2023} RoPE3D to incorporate spatial information of observations into the embedding, further enhancing the pattern separability in the memory when using bipolar data.

\subsection*{Contribution}

In this work, we present a hetero-associative sequential memory model that leverages neuromorphic signals for efficient storage and retrieval of robotic action sequences, and validate it on a service robot platform in various applications. In summary, the main contributions of this paper are:

\begin{itemize}
	\item We design a novel memory framework that binds robot joint states and tactile observations into compact bipolar binary vectors, enabling the storage of long sequential patterns in a memory-efficient form with minimal setup and training.
	\item We introduce a 3D rotary positional embedding (RoPE3D) that encodes tactile forces and spatial information in physical space, thereby enhancing separability in the binary vector space and enabling geometry-aware retrieval.
	\item We validate the proposed system on a service robot, where it realizes a pseudo-compliance controller that moves robot links proportionally to applied force, and accomplishes tactile-guided multi-joint grasp sequences recall through stepwise guidance.

\end{itemize}


\section{METHODOLOGY}

Consider $\boldsymbol{s}$ to be a vector of system state,  $\boldsymbol{S} = [\boldsymbol{s_1} \rightarrow \boldsymbol{s_2} \rightarrow \cdots \boldsymbol{s_l}] $ to be a system state sequence, $\mathbb{S} = [\boldsymbol{S_1}, \boldsymbol{S_2}, \ldots, \boldsymbol{S_i}]$ to be a collection of sequences. $\boldsymbol{o}$ is an observation vector. $\boldsymbol{O} = [\boldsymbol{o_1},  \boldsymbol{o_2}, \ldots, \boldsymbol{o_l}]$ is observations outside of the system, e.g., force measured by the sensor.  $\mathbb{O} = [\boldsymbol{O_1}, \boldsymbol{O_2}, \ldots, \boldsymbol{O_i}]$ is a collection of observations. The task of the hetero-associative sequential memory system (HASMS) is to make decision $\boldsymbol{d}$ when the system is in state $\boldsymbol{s}$ and has the observation $\boldsymbol{o}$ based on the hetero-associative sequential memory $\mathbb{M} = [\boldsymbol{M_1}, \boldsymbol{M_2}, \ldots, \boldsymbol{M_s}]$, where $\boldsymbol{M} = [\boldsymbol{m_1}, \boldsymbol{m_2}, \cdots, \boldsymbol{m_s} ]$ is a sequence of learned association $\boldsymbol{m}$ of $\boldsymbol{s}$ and $\boldsymbol{o}$. $\boldsymbol{D} = [\boldsymbol{d_1}, \boldsymbol{d_2}, \ldots, \boldsymbol{d_m}]$ denotes the decision under the current state and observation. $\mathbb{D} = [\boldsymbol{D_1}, \boldsymbol{D_2}, \ldots, \boldsymbol{D_i}]$ is a collection of decisions.  The task can be formulated as $\mathbb{D} = (\mathbb{S} \otimes \mathbb{O})\mathbb{M}$, where $\otimes$ denotes establishing a relationship between two variables, i.e., binding.

\subsection{Neuromorphic Encoding}

To mitigate the information annihilation issue, we first substitute $0$s in the neuromorphic data with $-1$s to construct the bipolar data.

The Rotary Positional Embedding 3D (RoPE3D) converts bipolar data of d-dimensional space into d/3 sub-spaces, and rotates the sub-spaces by the 3D rotations described by equations (\ref{rope3d1}) and (\ref{rope3d2}). 

\begin{equation}
\label{rope3d1}
RoPE3D(\mathbf{q}) = \begin{pmatrix}
\mathbf{R}_1 & & & \\
& \mathbf{R}_2 & & \\
& & \ddots & \\
& & & \mathbf{R}_{d/3}
\end{pmatrix} \begin{pmatrix}
q_1 \\
q_2 \\
\vdots \\
q_d
\end{pmatrix}, 
\end{equation}

\begin{equation}
\label{rope3d2}
\begin{aligned}
\mathbf{R}(\mathbf{n}, \theta) 
&= \mathbf{I} + \sin(\omega_i \theta) \, [\mathbf{n}]_{\times} 
   + (1-\cos(\omega_i \theta)) \, [\mathbf{n}]_{\times}^2 , \\[6pt]
\omega_i
&=  10^{- \tfrac{i}{d}}, 
   \quad i \in \{1, 2, \ldots, \tfrac{d}{3}\} ,
\end{aligned}
\end{equation}

where $\mathbf{R_i}$ is the rotation matrix in 3D space, $\mathbf{n}$ the 3D vector formed by every three elements of $\mathbf{q}$. $\theta$ is the rotation angle corresponding to observations (e.g., the angle of incoming forces from sensory input). $\omega_i$ denotes the frequency scaling of the rotation angle $\theta$ varying with the positional index of the vector $i$.  $[{n}]_{\times}$ is the skew-symmetric matrix of axis of rotation $\mathbf{n}$.
\begin{equation}
	[\mathbf{n}]_{\times} = \begin{pmatrix}
0 & -n_z & n_y \\
n_z & 0 & -n_x \\
-n_y & n_x & 0
\end{pmatrix}.
\end{equation}

\subsection{Association Creation}
For the binding of two vectors, elementwise multiplication provides a computationally straightforward and theoretically well-founded approach, and also preserves the similarity of the original vectors \cite{yuUnderstandingHyperdimensionalComputing}.

After binary data encoding, the bipolar encoded system state $\boldsymbol{s}_{encoded}$ and RoPE3D encoded observation $\boldsymbol{o}_{encoded}$ are binded to create association $\boldsymbol{m}$
\begin{equation}
	\boldsymbol{m} = \boldsymbol{s}_{encoded} \otimes \boldsymbol{o}_{encoded}.
\end{equation}

\subsection{Inference}
The association of Sequences $\mathbb{S}$ and observations $\mathbb{O}$ is stored in the association matrix $\mathbb{M}$. During the inference time, the new joint states $\boldsymbol{s}$ and new observation $\boldsymbol{o}$ are encoded into the same space as the association, and a query vector $\boldsymbol{q} = \boldsymbol{s}\_encoded  \otimes\boldsymbol{o}\_encoded$ is employed to make decision of new system state $\boldsymbol{d}$ from the memory matrix $\mathbb{M}$. Thanks to the softmax operation, the new decisions can be formulated by a weighted combination of multiple possible memories that correlated.

\subsection{Hetero-associative Sequential Memory}

The update rule of the memory is
\begin{align}
\boldsymbol{\boldsymbol{d}} 
    &= f(\boldsymbol{\boldsymbol{s_c}}) \\
    &= \mathbb{S}_{-1} \,\text{softmax}(\beta \mathbb{M}^T \boldsymbol{q}) ,
\end{align}
where $\boldsymbol{\boldsymbol{q}}$ is a query vector that represents the association of current state and current observation, $\mathbb{M}$ is the associative memory, $\mathbb{S}_{-1}$ is the back-shifted matrix of action patterns for a sequential recall, $\beta$ denotes the association scaling factor, controlling the association fuzziness of the inference \cite{krotovDenseAssociativeMemory2016}. A value of $\beta$ close to 0 makes the associative memory preferentially infer the main feature of the stored system state patterns, while a value bigger than 1 makes the memory choose the specific stored system state pattern.

%
%

\section{EXPERIMENT}
\subsection{Hardware Platform}

\begin{figure}[h] 
    \centering
        \includegraphics[width=0.45\textwidth]{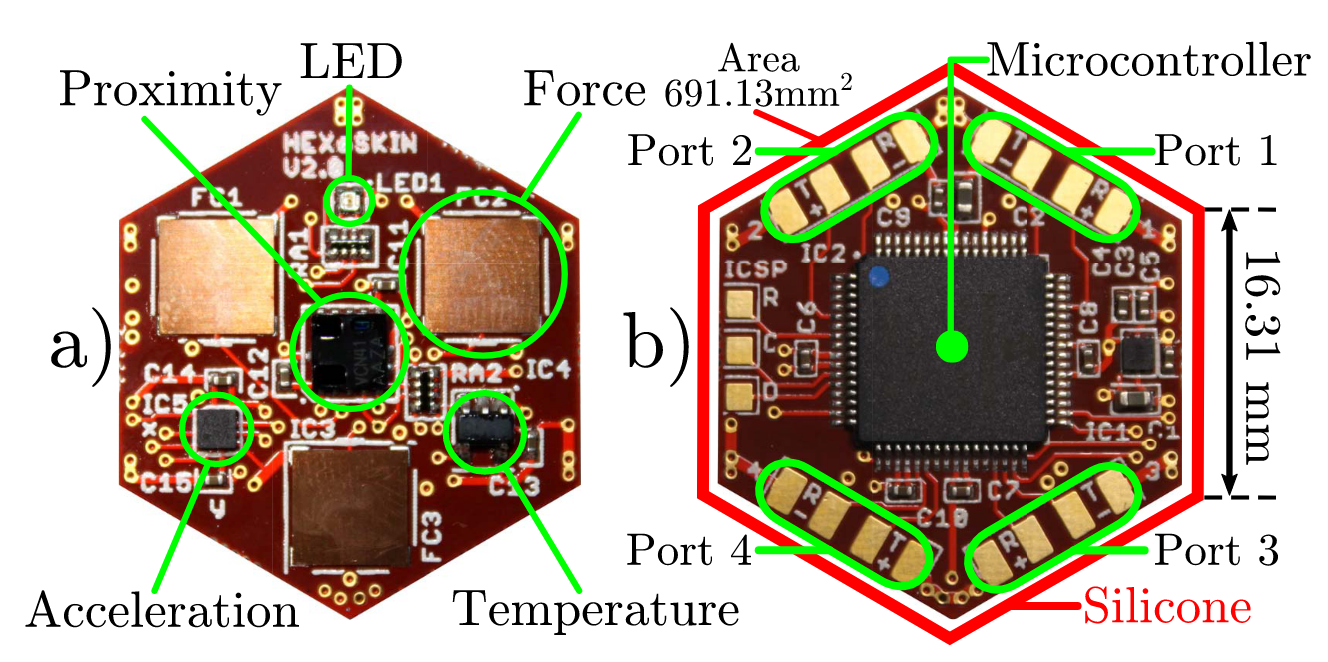} 
        \caption{Robot skin cell. (a) Sensor distribution on the cell. (b) Microcontroller, connectivity, and dimensions of the cell.}
        \label{fig: skin_inside}
\end{figure}

\begin{figure}[h] 
    \centering
        \includegraphics[width=0.3\textwidth, angle=-90]{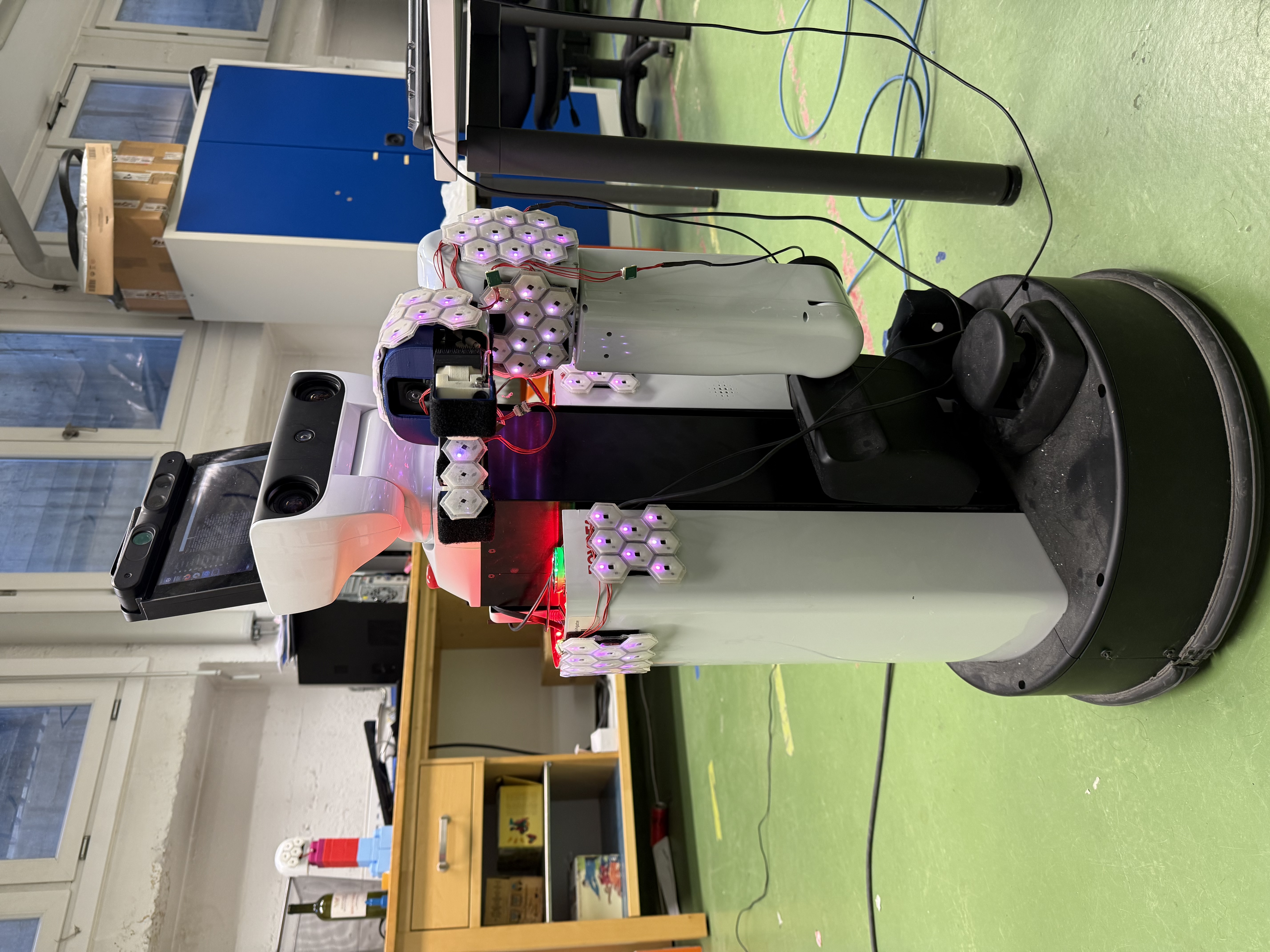} 
        \caption{Illustration of the HSR, whose hand, wrist, and part of the body are covered with robot skin cells.}
        \label{fig: hsr}
\end{figure}

In the experiment, the robot we use is the Toyota Human Support Robot (HSR), which is designed to provide support and assistance in home and healthcare environments \cite{ToyotaHumanSupport2021}. The HSR is partially covered with robot skin \cite{chengComprehensiveRealizationRobot2019} as illustrated in Fig. \ref{fig: hsr} to sense external forces. Fig. \ref{fig: skin_inside} shows the structure of a skin cell. A skin patch consists of multiple skin cells, and it also provides the spatial information of each cell, enabling precise localization of external stimuli. Specifically, we mounted four skin patches on every lateral side of the wrist, and three patches on the torso.


\subsection{Robotic Application}
To validate the capability of the memory model, we designed the following two applications:
\begin{itemize}
	\item Pseudo-compliance control: Develop a controller that allows the robot links to move in the direction of the force detected by the robot skin, with velocity positively correlated to the force amplitude.
	\item Tactile-guided grasp execution: Enable the robot to perform step-by-step grasp actions under continuous human tactile guidance.
\end{itemize}

In this experiment, HSR provides the joint angles as the system state $\boldsymbol{s}$. During the execution of an action, the sequence of joint angles of the robot forms the system state sequence $\boldsymbol{S} = [\boldsymbol{s_1} \rightarrow \boldsymbol{s_2} \rightarrow \cdots \boldsymbol{s_l}]$. The system state sequence obtained in the course of data collection constitutes $\mathbb{S} = [\boldsymbol{S_1}, \boldsymbol{S_2}, \ldots, \boldsymbol{S_i}]$.

Similarly, $\boldsymbol{o}$ is the force signal sensed by skin patches. $\boldsymbol{O} = [\boldsymbol{o_1},  \boldsymbol{o_2}, \ldots, \boldsymbol{o_l}]$ is a collection of skin signals synchronized with $\boldsymbol{s}$. $\mathbb{O} = [\boldsymbol{O_1}, \boldsymbol{O_2}, \ldots, \boldsymbol{O_i}]$ denotes force signals corresponding to multiple sequences.

 As consequence, $\boldsymbol{m}$ is the association of the joint state $\boldsymbol{s}$ and corresponding force $\boldsymbol{o}$, $\boldsymbol{M} = [\boldsymbol{m_1}, \boldsymbol{m_2}, \cdots, \boldsymbol{m_s} ]$ is the hetero-associative sequential memory of one sequence.
The desired joint state is provided by the output of the hetero-associative memory model given the current state $\boldsymbol{s}$ and observation $\boldsymbol{o}$. The robot serves as the agent responsible for decisions made by the hetero-associative memory.

\subsection{Population Place Coding of Joint Angles}

To encode the angle of a single joint, we employed the bio-plausible population place coding method \cite{georgopoulosNeuronalPopulationCoding1986}. We use a group of $N_p=10$ neurons with different preferred angles to represent the joint angle with a collective response pattern of the population. The neuronal responses follow a Gaussian tuning curve, defined as:
\begin{equation}
		r^j_i = \exp\left(-\frac{(\phi^j - \phi^j_i)^2}{2\sigma^2}\right),
		\label{eq:tuning_curve}
\end{equation}
where: 
\begin{itemize}
	\item $r^j_i$ is the spike rate density of the $i$-th neuron that encode the $j$-th joint,
	\item $\phi$ is the measured joint angle,
	\item $\phi^j_i$ is the preferred angle of neuron $i$ that is evenly distributed across the joint limit,
	\item $\sigma$ is the tuning width parameter.
\end{itemize}
$r^j_i$ is quantized to a binary joint place code vector $\textbf{p}_j$ by $n$ bit-shift, keeping only the most significant $n$ bits. We denote the number of joints as $N_J$. The global place code vector $\textbf{P}$ of length $n \cdot N_j$ is constructed by concatenating all the joint place code vectors.




\subsection{Neuroromorphic Encoding of External Forces}
\subsubsection{Preprocessing}
For a cell $c$, the normal force applied on the cell is measured by three force sensors as illustrated in Fig. \ref{fig: skin_inside}:

\[
\mathbf{f}_c = \bigl(f_{c,1},\, f_{c,2},\, f_{c,3}\bigr),
\]

where $f_{c,1}, f_{c,2}, f_{c,3}$ denote the forces measured by three force sensors, respectively.

The total force magnitude on the skin cell $c$ is calculated through the L1 norm:

\[
\|\mathbf{f}_c\|_{1} \;=\; |f_{c,1}| + |f_{c,2}| + |f_{c,3}|.
\]

A soft-thresholding operation is subsequently applied to attenuate the noise of small magnitude. For a given threshold $\tau$, the effective contribution of cell $c$ at time $t$ is:

\[
m_c(t) = \max\bigl(0, \|\mathbf{f}_c\|_{1} - \tau \bigr).
\]

Finally, the total effective force at time $t_0$ is obtained by accumulating the contributions within a temporal window of length $\Delta t$:

\[
F_{\text{total}} = \sum_{t\in [t_0-\Delta t,t_0]} m_c(t).
\]


\subsubsection{Izhikevich Neuron Coding}
After preprocessing, the force signals measured by the robot skin patches are encoded into spike trains with the bio-plausible Izhikevich neuron model \cite{izhikevichSimpleModelSpiking2003}. The internal states are described as follows:

\begin{align}
\label{eq:izhi_1}
\frac{dv_i}{dt} &= 0.04v_i^2 + 5v_i + 140 - u_i + I \\
\label{eq:izhi_2}
\frac{du_i}{dt} &= a(bv_i - u_i)\\
\text{if } v_i &\geq v_{\text{th}}, \text{ then }
\begin{cases}
v_i \leftarrow c \\
u_i \leftarrow u_i + d
\end{cases}
\end{align}
where $v$ represents the membrane potential of the neuron, $u$ is a membrane recovery variable, and $I$ is synaptic currents, which is proportional to $F_{total}$.
	
The cell emits a spike when the membrane potential reaches its threshold. 
The parameters are adjusted such that the neuron response is primarily related to the amplitude of force as listed in the table \ref{tab:izhikevich_params}. 

\begin{table}[htbp]
\centering
\caption{Parameters for Izhikevich Neuron Model}
\begin{tabular}{|c|c|p{4cm}|c|}
\hline
\textbf{Parameter} & \textbf{Default Value} & \textbf{Description}  \\
\hline
$a$ & 0.02 & Recovery time scale  \\
\hline
$b$ & 0.20 & Sensitivity of recovery variable  \\
\hline
$c$ & -50.0 & Post-spike reset value for $v$ \\
\hline
$d$ & 0.50 & Post-spike reset value adjustment \\
\hline
$v_{th}$ & 30 & Post-spike reset value adjustment \\
\hline
\end{tabular}
\label{tab:izhikevich_params}
\end{table}


\subsection{Tactile Embedding}

We extract the spike rate density $\rho$ of the Izhikevich neuron models as a force feature and employ it for the tactile embedding:
\[
  \rho = \clamp\!\bigl(\tfrac{r_f}{R},\, 0,\, 1\bigr).
\]
where $r_f$ is the measured spike rate of an Izhikevich neuron and $R$ is a parameter that represents the maximum spike rate.

 The bipolar force embedding vector $F \in \{-1,+1\}^{N_n \cdot N_J}$ is initilized with all entries set to $-1$. The indices corresponding to the largest $N=\round(\rho \cdot N_n \cdot N_J)$ values in the global place code vector $\textbf{P}$ are identified. The entries in $F$ at these indices are flipped to $+1$.
 
To bind this force to the binary joint representation, we expand $F$ by repeating each entry $n$ times. This yields an expanded bipolar vector of length $(n \cdot N_n \cdot N_J)$, matching the dimensionality of the encoded joint state.

The expanded force vector is then passed through the \texttt{RoPE3D} to obtain the final tactile representation for learning and inference. During learning, each joint-state sample is temporally paired with the closest spike within a small time window, and associations are formed between the encoded joint state and the encoded force. 

\subsection{Data Collection}

 In the data collection process, the applied force inputs and the corresponding desired movement are recorded synchronously by pairing joint states with the closest force signal within a narrow temporal window. The sequence of force magnitude, together with the rotation information of the skin patch, is embedded into tactile embedding $\boldsymbol{O}$ using RoPE3D, after which it is bound to the corresponding joint state through the operation $\boldsymbol{S} \otimes \boldsymbol{O}$, where $\boldsymbol{S}$ denotes the neural representation of joint states. The resulting association, denoted by $\boldsymbol{m}$, is stored in the memory matrix $\mathbb{M}$, which serves as the repository of learned joint-force correspondences. In parallel, the action sequences observed during data collection are stored in a temporally shifted form within $\boldsymbol{S_{-1}}$ to align present associations with the appropriate future joint states in the inference.

\subsection{Inference}
During the inference, the observation $\boldsymbol{o}$, i.e., the tactile embedding, is generated through RoPE3D using the rotation information of the skin patch and the sensed external force magnitude. $\boldsymbol{o}$ is bound with the encoded joint state $\boldsymbol{s}$ to form the query vector $\boldsymbol{q} = \boldsymbol{s} \otimes \boldsymbol{o}$. The query is then matched against the memory matrix $\mathbb{M}$, which stores previously learned associations. Similarity scores are computed between $\boldsymbol{q}$ and the stored associations, through softmax, the new decision $\boldsymbol{d}$ is obtained by a weighted combination of the corresponding value entries from $\boldsymbol{S_{-1}}$. The retrieved result $\boldsymbol{d}$ represents a target joint state, which is used to generate the appropriate motor command.

\section{Result}

\subsubsection{Pseudo-compliance Controller}
\begin{figure}[htbp]
  \centering

  \subfloat[The arm moves upward 1]{\includegraphics[width=0.2\textwidth]{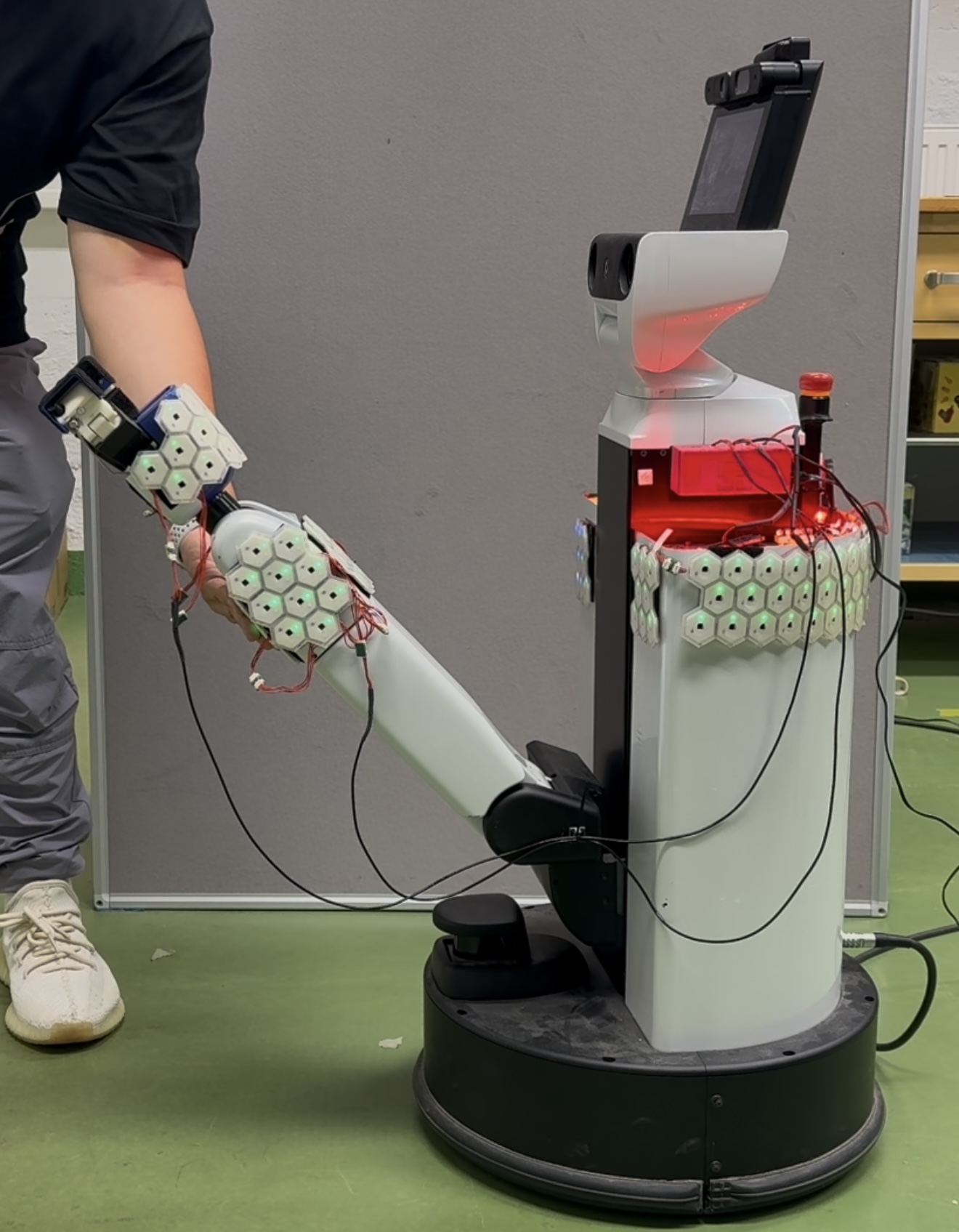}}\hfill
  \subfloat[The arm moves upward 2]{\includegraphics[width=0.2\textwidth]{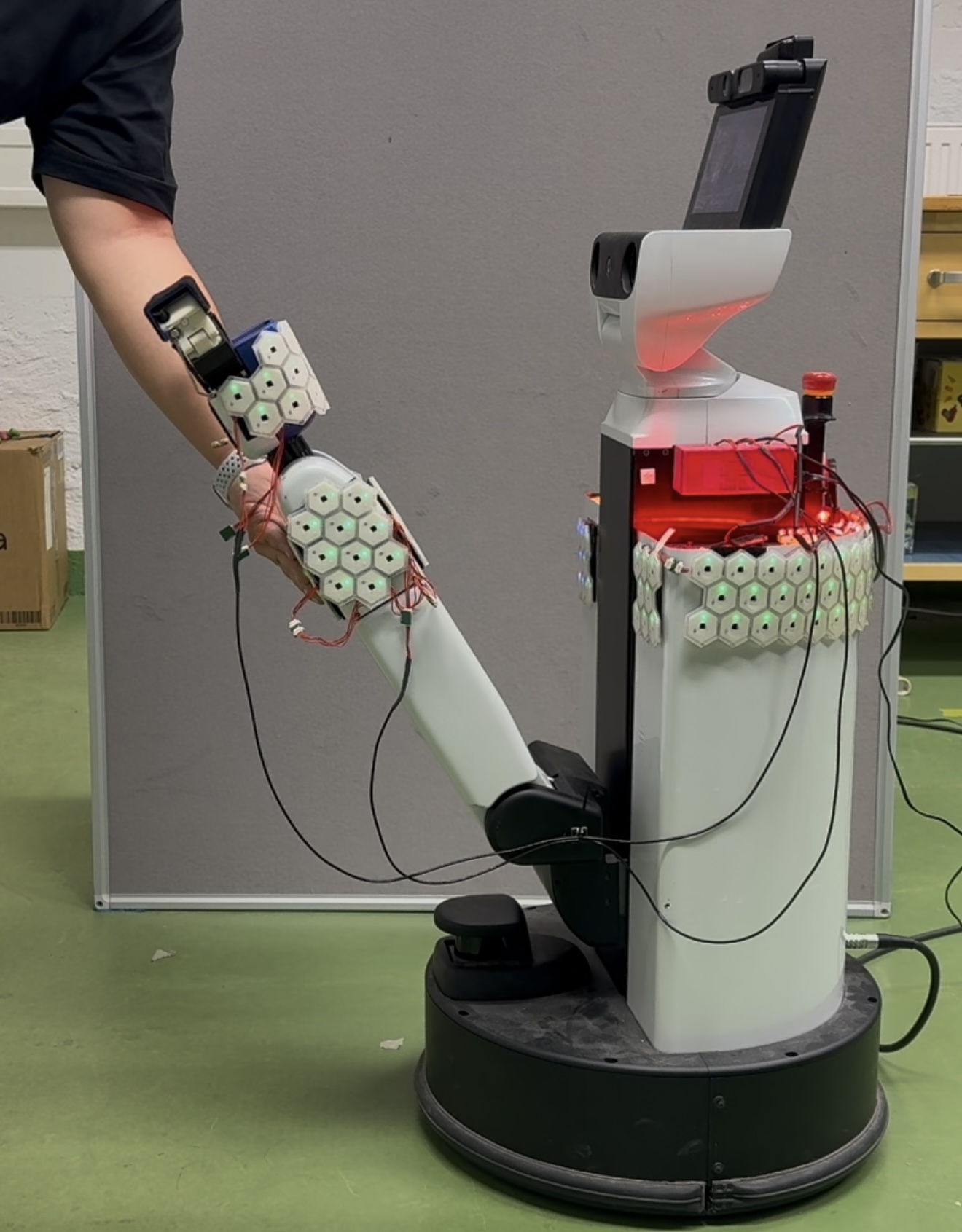}} 

  \subfloat[The arm moves downward 1]{\includegraphics[width=0.2\textwidth]{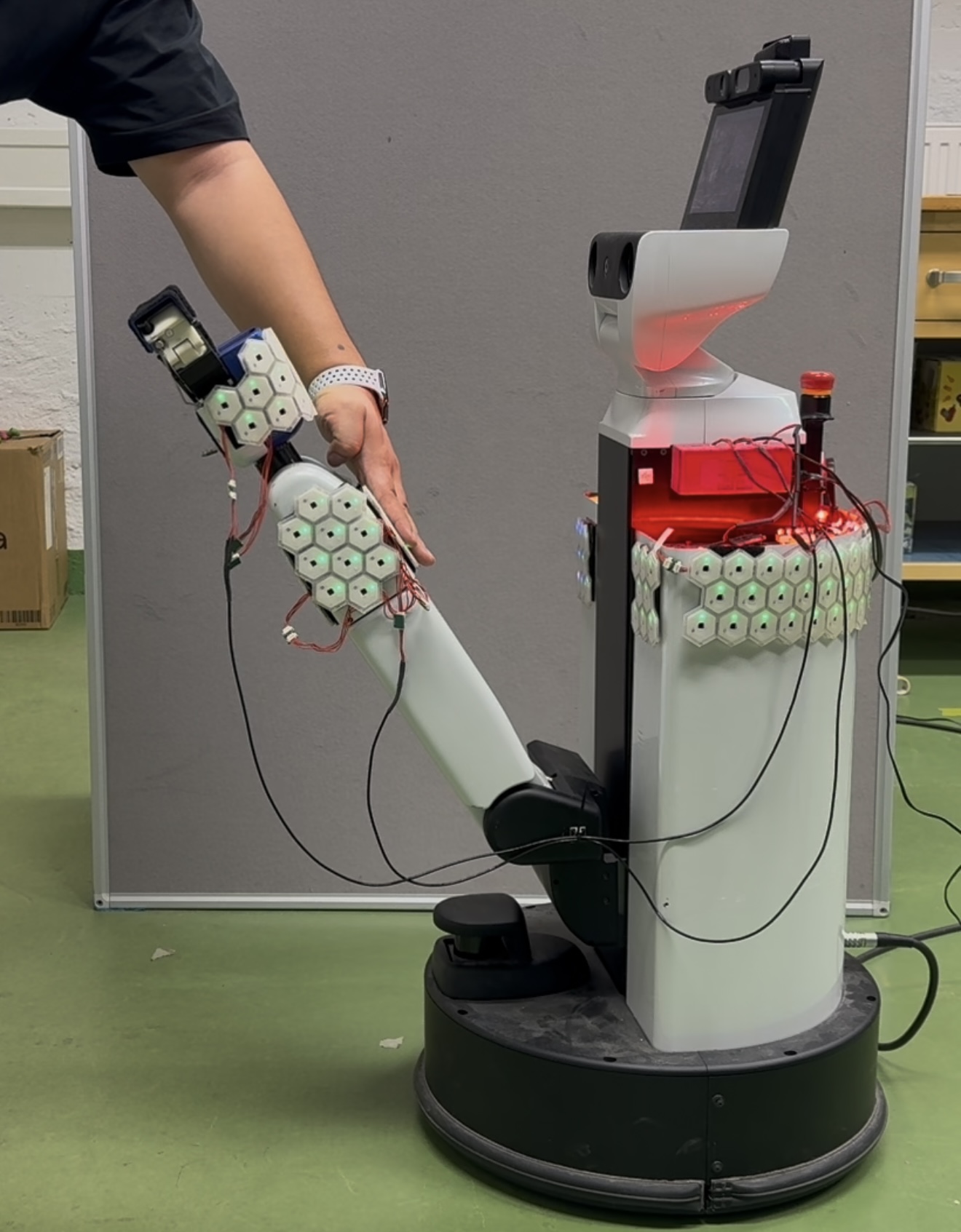}}\hfill
  \subfloat[The arm moves downward 2]{\includegraphics[width=0.2\textwidth]{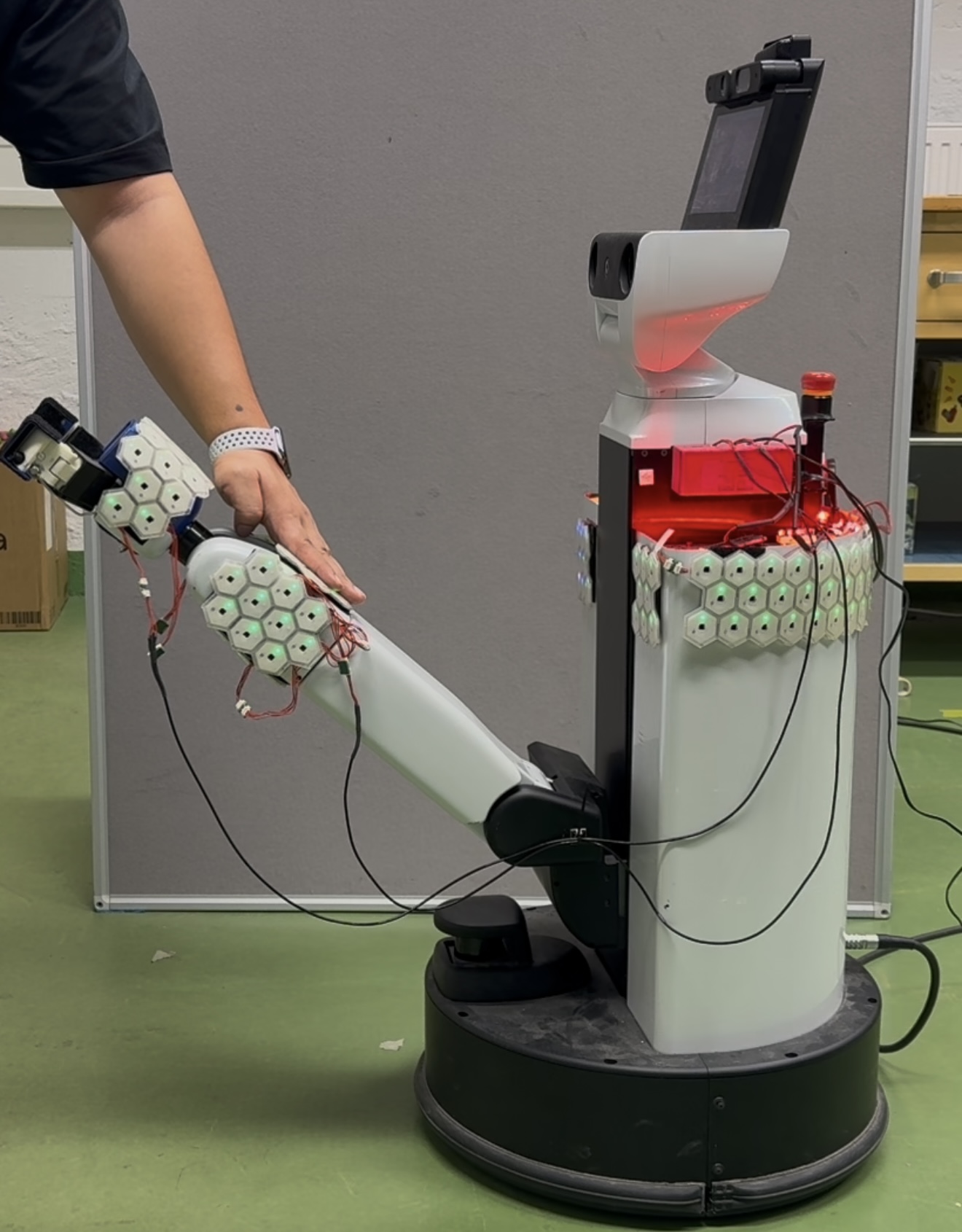}} \\

  \subfloat[The wrist rotates anti-clockwise 1]{\includegraphics[width=0.2\textwidth]{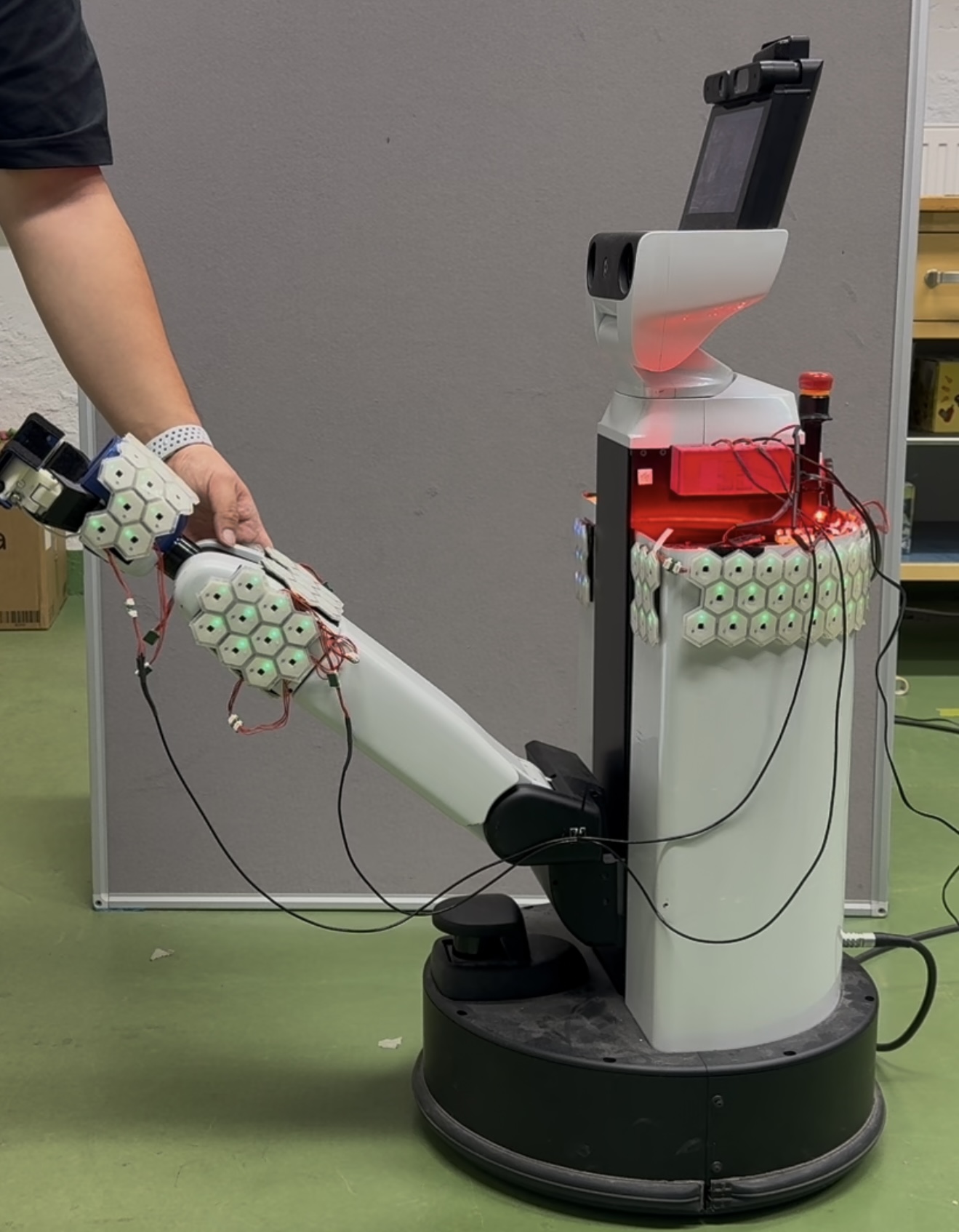}}\hfill
  \subfloat[The arm rotates anti-clockwise 2]{\includegraphics[width=0.2\textwidth]{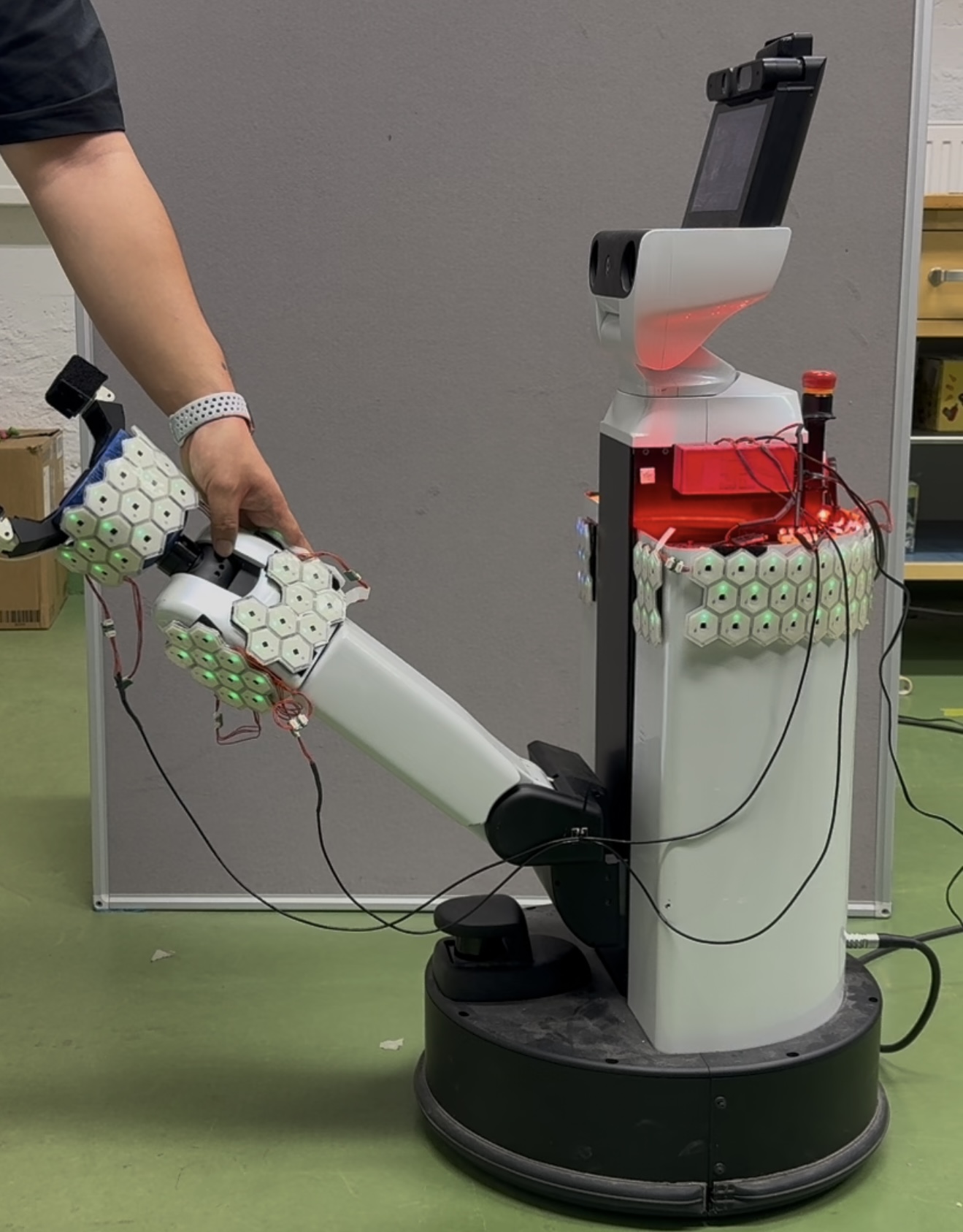}} \\

  \subfloat[The wrist rotates clockwise 1]{\includegraphics[width=0.2\textwidth]{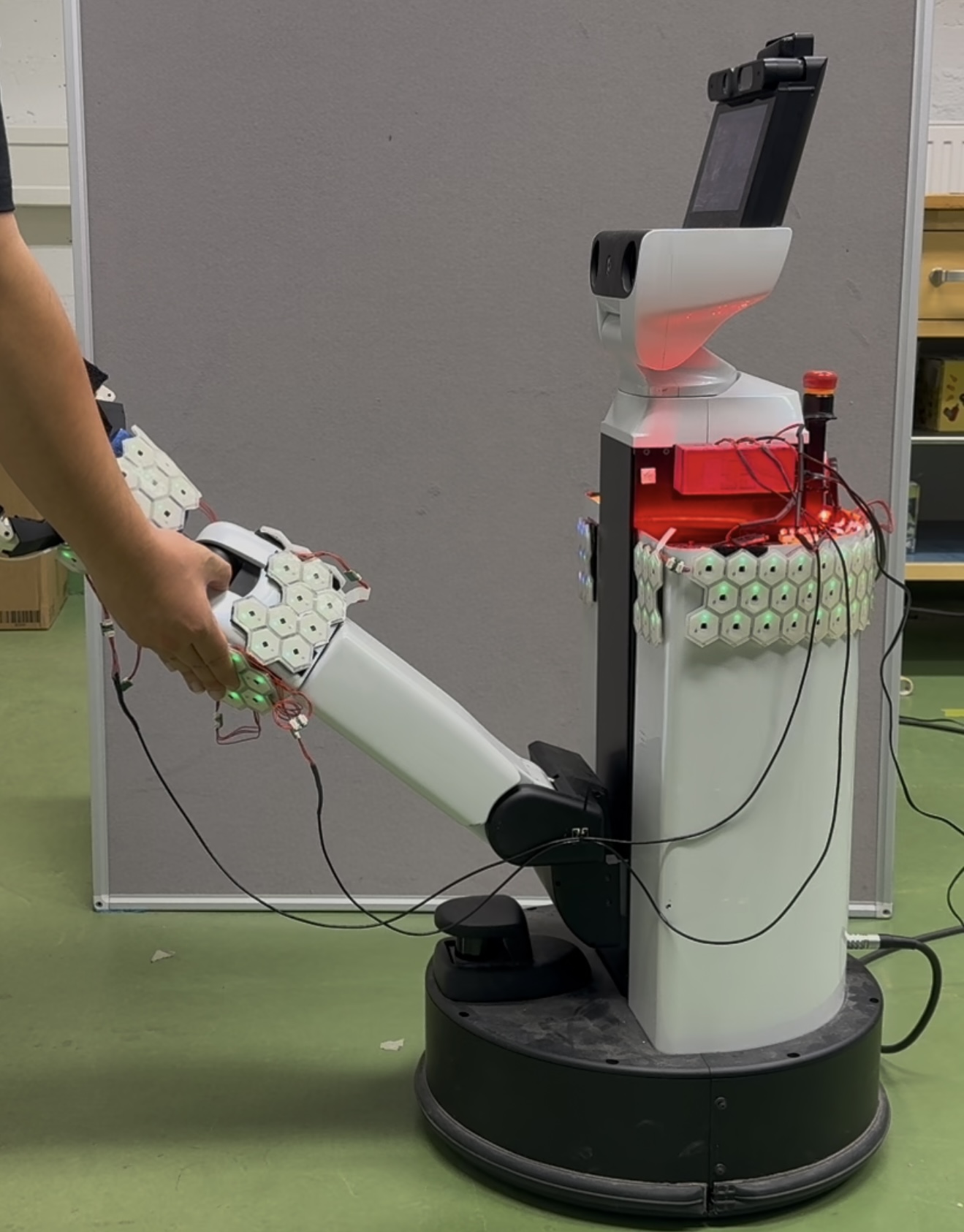}}\hfill
  \subfloat[The arm rotates clockwise 2]{\includegraphics[width=0.2\textwidth]{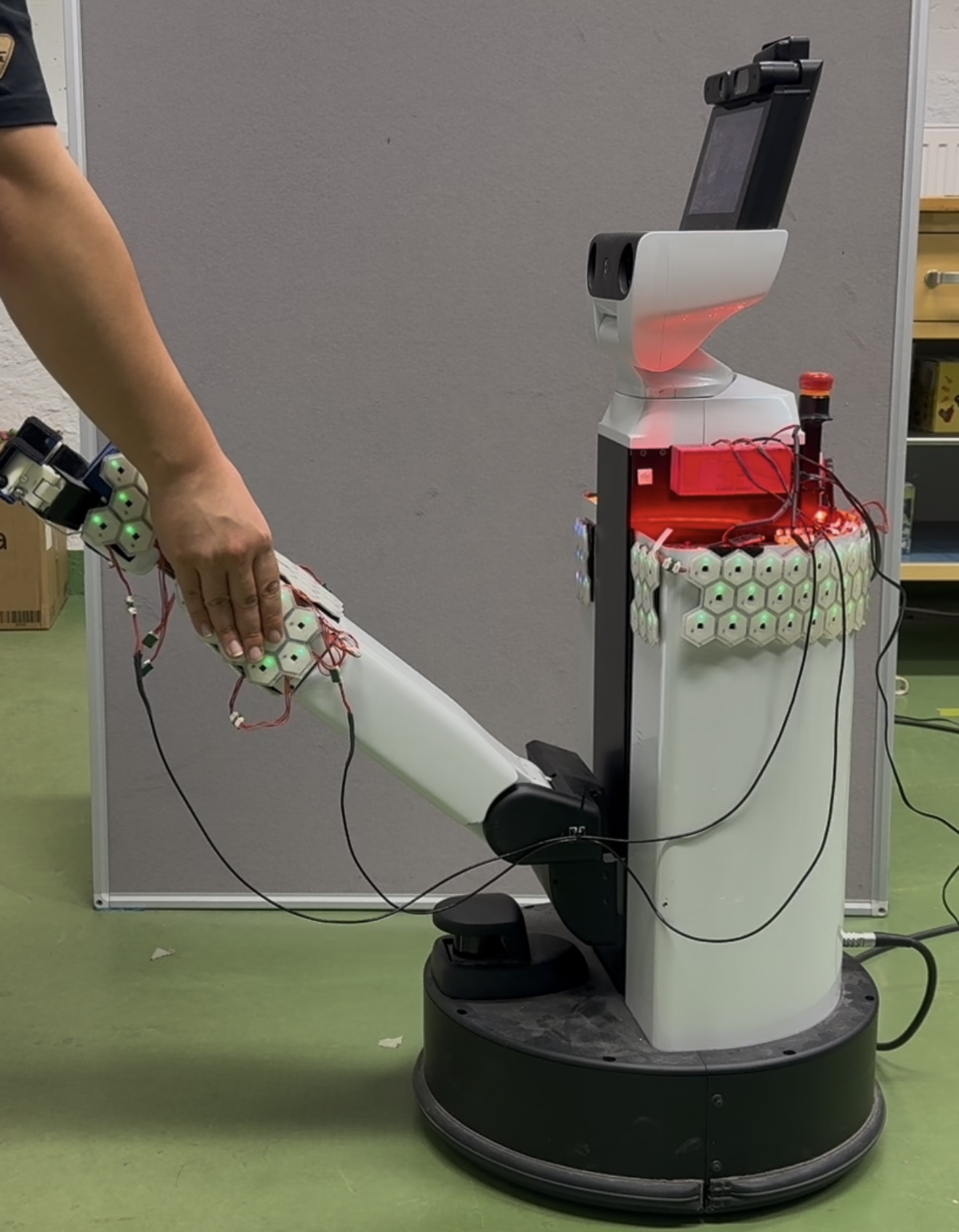}}

  \caption{Demonstration of the pseudo-compliance controller: the robotic arm is guided by gentle touches on skin patches, enabling intuitive human interaction}
  \label{fig:compliance}
\end{figure}

\begin{figure*}[htbp]
    \centering
        \includegraphics[width=1\textwidth]{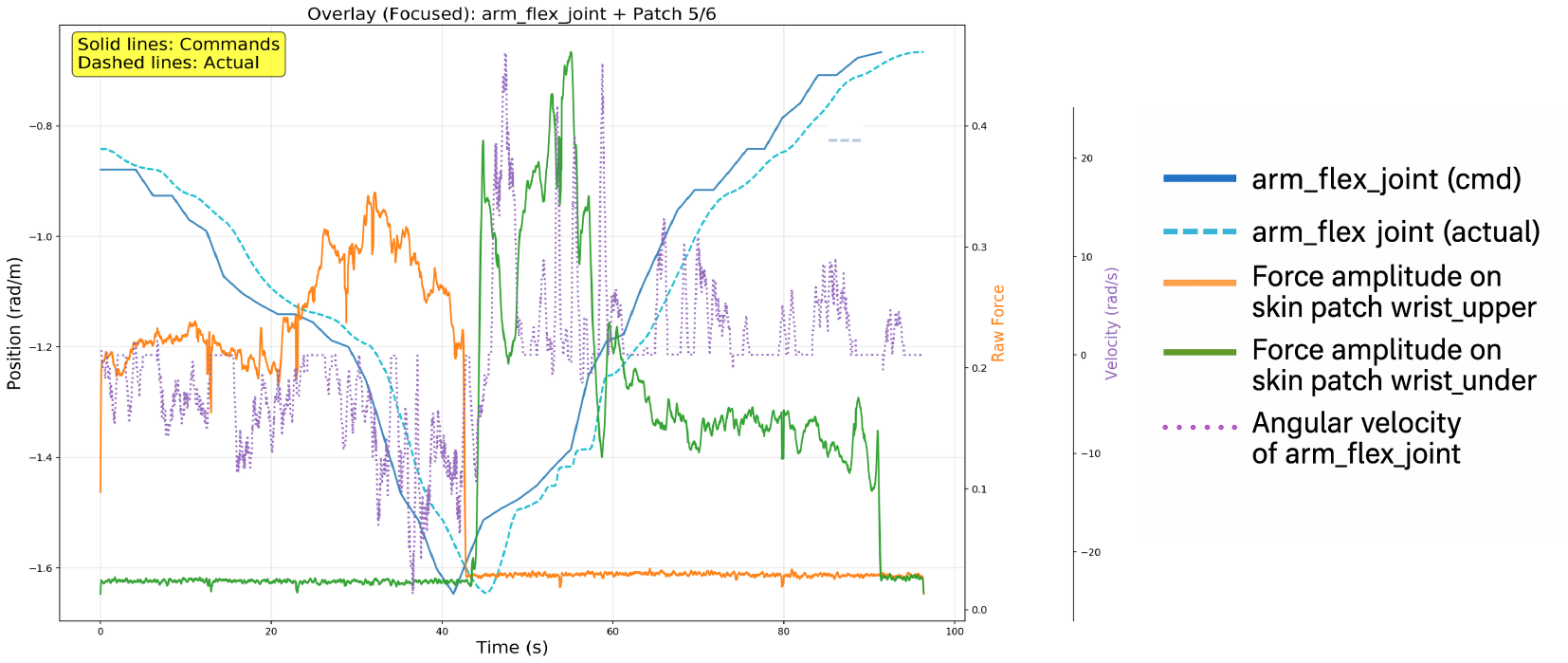} 
        \caption{Trajectory, force amplitude, and joint angular velocity of arm\_flex\_joint of the HSR under the pseudo-compliance control. The forces on the skin patch wrist\_upper and wrist\_under are opposite forces. The joint angular velocity is positively correlated with the force applied to the skin patches.}
        \label{fig:compliance_traj}
\end{figure*}

With the memory model, a pseudo-compliance controller can be realized by simply allowing the model to memorize the complete unidirectional trajectory of a single joint within the joint range and corresponding temporally aligned interaction. In comparison to the analytic compliance controller, it does not require a precise robot dynamic model, but only has to record the whole range of motion of the joint. The Fig. \ref{fig:compliance_traj} illustrates the joint trajectory, force amplitude, and joint angular velocity of the pseudo-compliance control on the robot arm. The arm is first pushed down by touching the wrist\_upper skin patch, then rises up with the opposite skin interaction. The force direction on the skin patch wrist\_upper and wrist\_under is opposite. With the increased force, the joint angular velocity increases in a positive correlation, which realizes compliance. For compliance in multiple joints and directions, we only need to create dedicated memory for each skin patch. Fig. \ref{fig:compliance} demonstrates the compliance behavior of the robot arm with four directional movements, which move upward, downward, or rotate in response to the applied force.

%
%
%
%

\subsubsection{Multi-joint Action Sequence}
\begin{figure}[htbp]
  \centering

  \subfloat[Seq 1]{\includegraphics[width=0.23\textwidth]{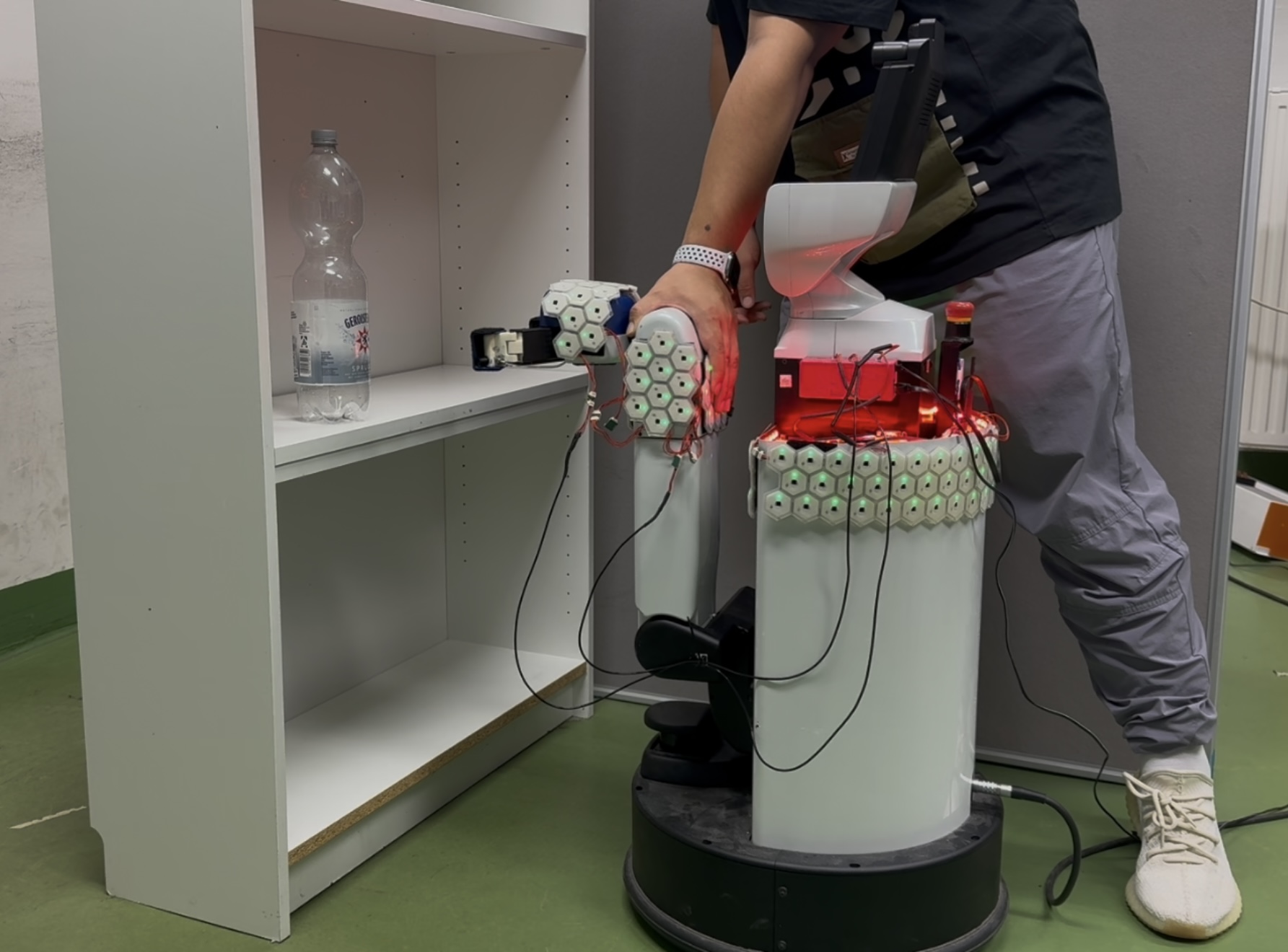}}\hfill
  \subfloat[Seq 2]{\includegraphics[width=0.23\textwidth]{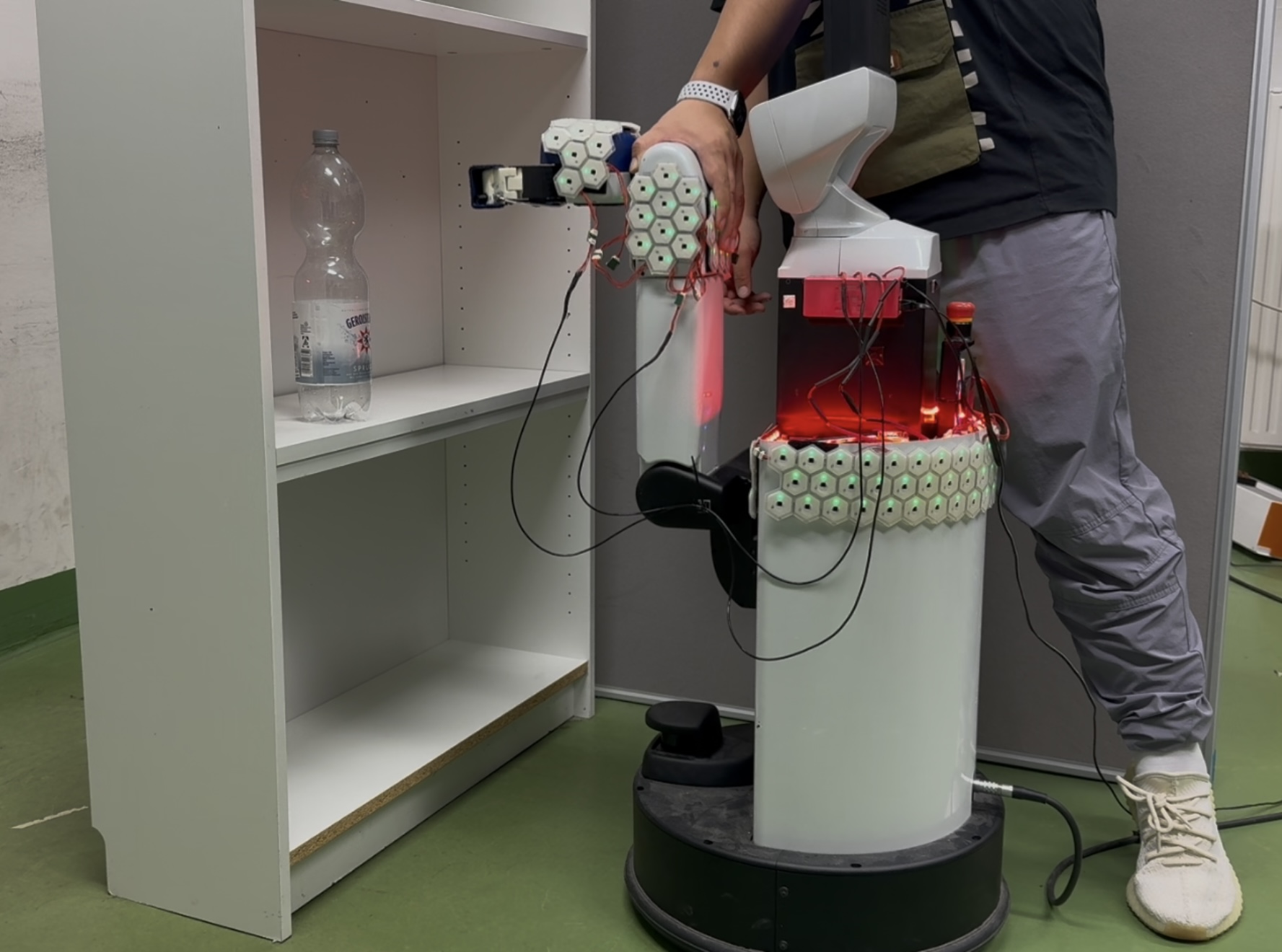}}\hfill
  
  \subfloat[Seq 3]{\includegraphics[width=0.23\textwidth]{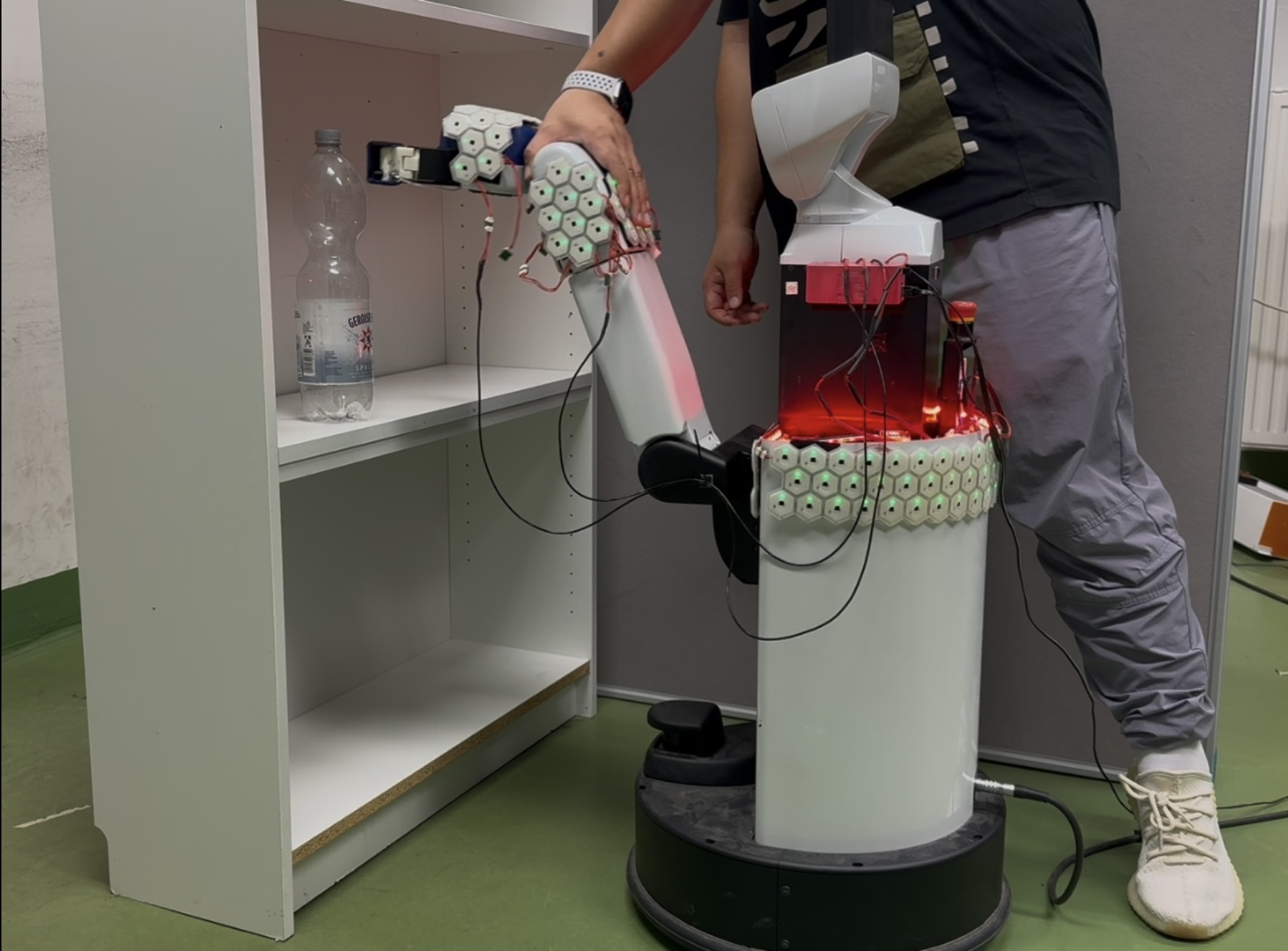}}\hfill
   \subfloat[Seq 4]{\includegraphics[width=0.23\textwidth]{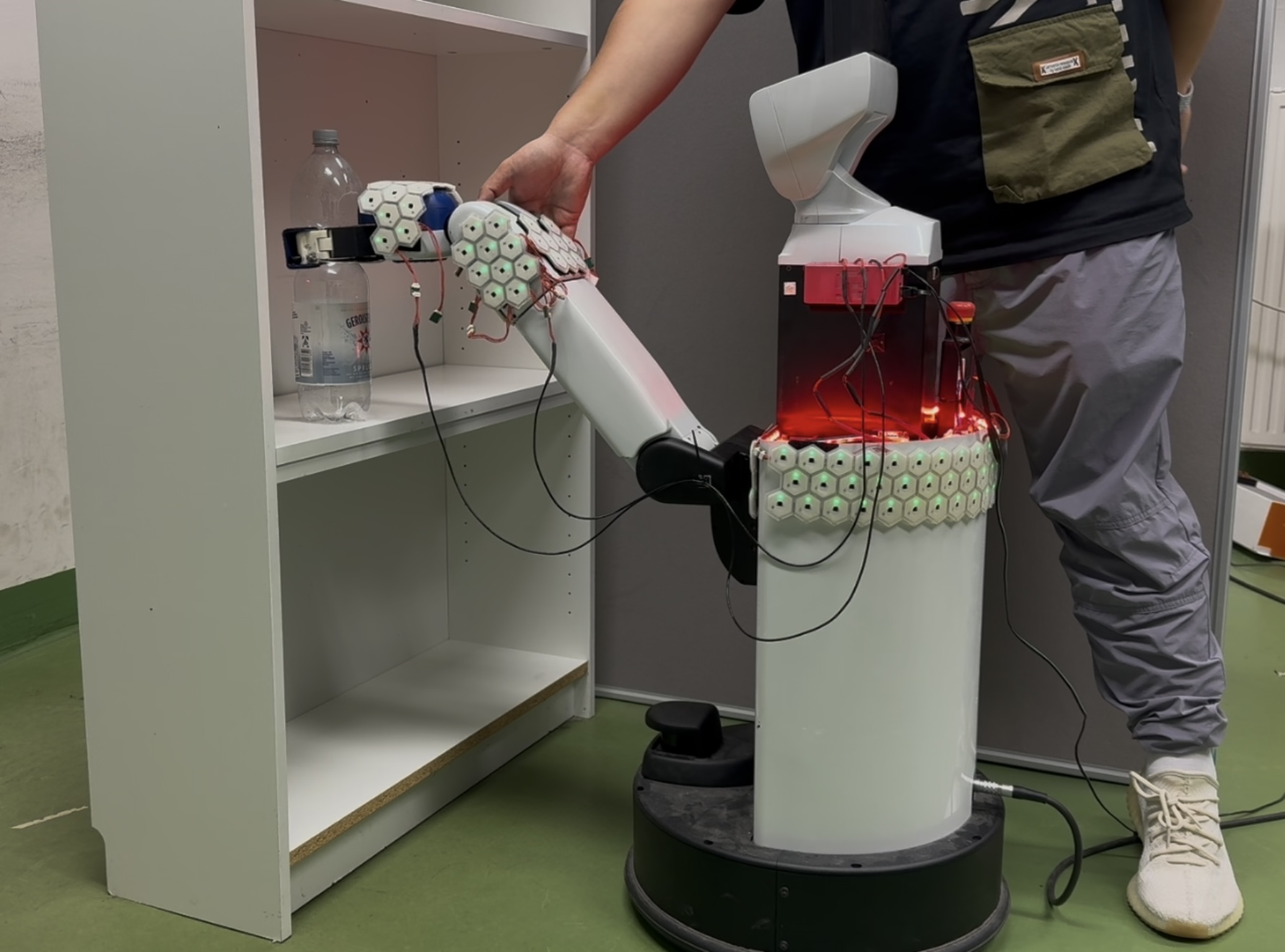}}\hfill
   
  \subfloat[Seq 5]{\includegraphics[width=0.23\textwidth]{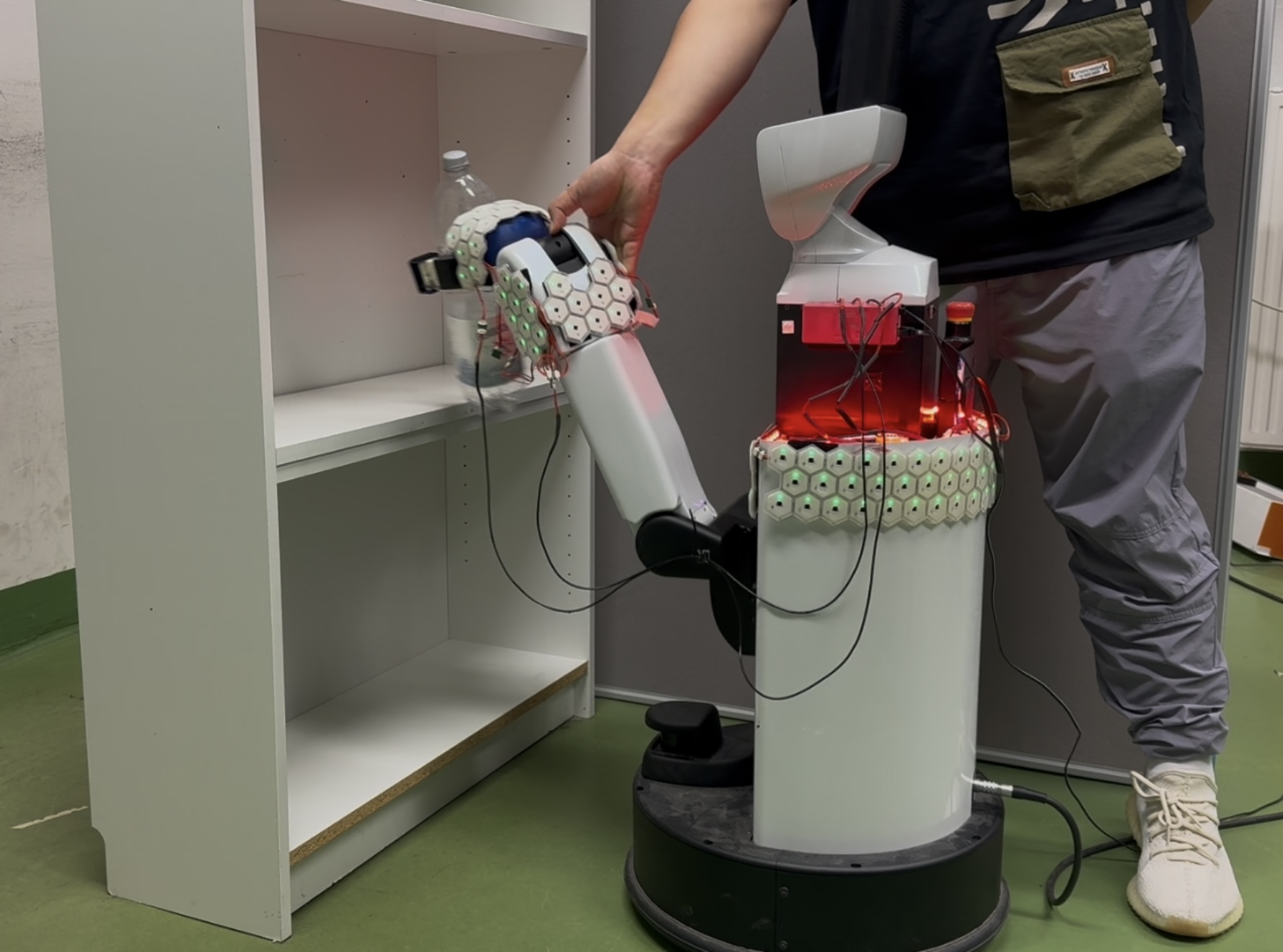}}\hfill
  \subfloat[Seq 6]{\includegraphics[width=0.23\textwidth]{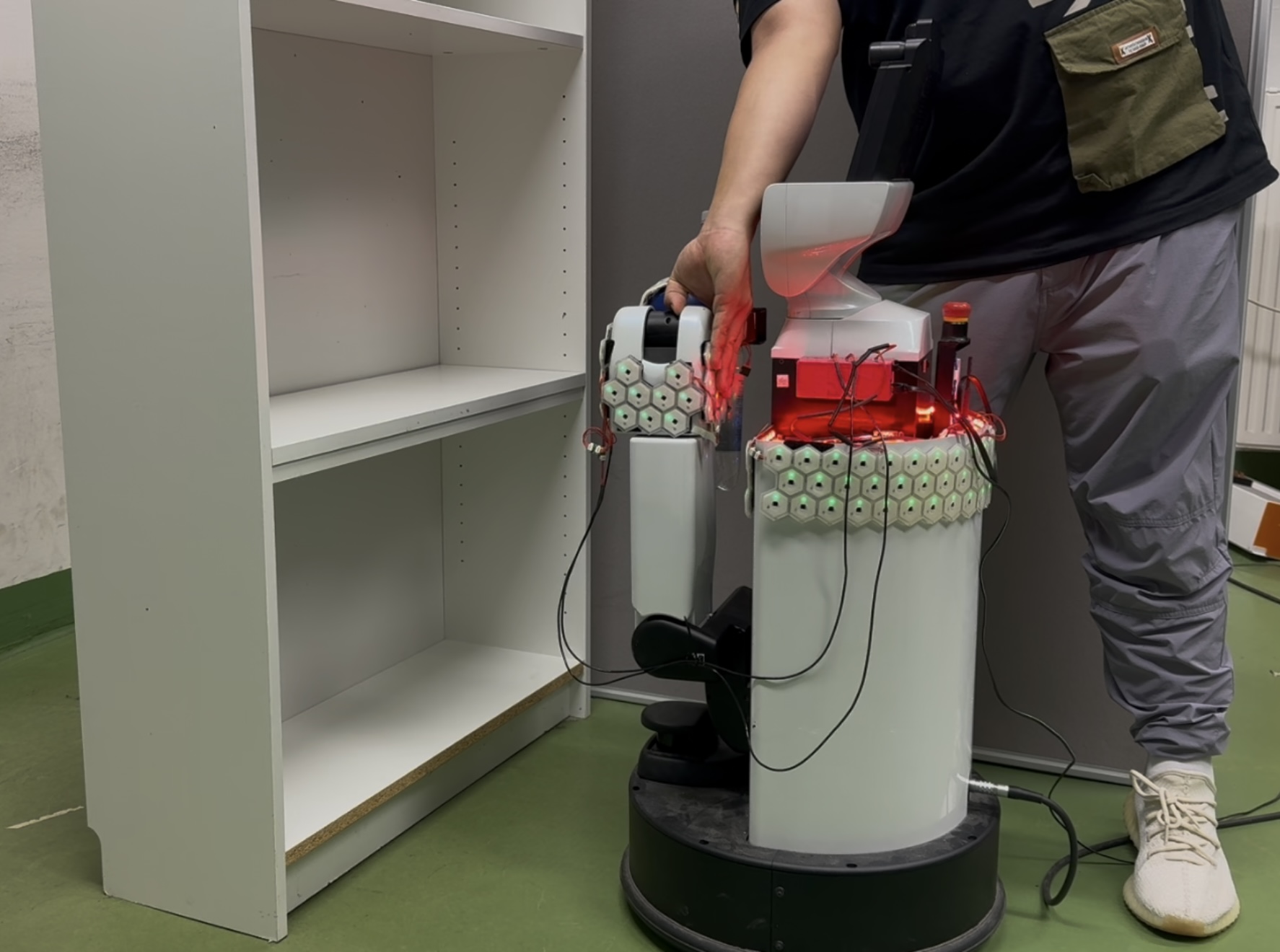}}\hfill

  \caption{Tactile-guided grasp execution: the robot performs step-by-step grasping actions under continuous human tactile input. In the first three figures, the robot reaches the item by following touch guidance on a single patch, while in the last three figures, it grasps the item and retracts the gripper in response to touch on a different patch.}
  \label{fig:sequence}
\end{figure}

The memory model is capable of storing and recalling complete sequences of trajectories across multiple joints. Under continuous tactile guidance, the robot executes the stored grasping actions step by step. Dedicated memories can be assigned to specific skin patches, allowing each patch to trigger distinct functionalities. As shown in Fig. \ref{fig:sequence}, the wrist\_upper patch is associated with reaching movements that position the gripper for grasping, while the wrist\_right patch is associated with grasping and subsequent retraction. In this way, the robot first reaches toward the object under tactile guidance, then completes the grasp by coordinating the torso, head, and arm joints, and finally retracts its hand in response to input on a different skin patch. This demonstrates the ability of the memory to perform coordinated multi-joint actions through associative recall.

%
%
%
%
%

\section{CONCLUSIONS}
In this work, we introduced a hetero-associative sequential memory model that leverages neuromorphic signals to enable efficient and scalable pattern storage for robotic applications. By encoding joint states with population place coding and tactile forces with an Izhikevich neuron model, and further enhancing separability through the proposed 3D rotary positional embedding, the system achieves robust associative recall in high-dimensional binary space. Experiments on a service robot demonstrated two representative applications: a pseudo-compliance controller that adapts robot motion to tactile input and a tactile-guided multi-joint grasp execution. The introduced model is easy to set up, economical in computation and memory usage, and exhibits a degree of generalization beyond simple memorization. These results suggest that hetero-associative sequential memory provides a promising alternative paradigm for motion control and action planning, while also opening future avenues for integration with imitation learning, multi-modal perception, and neuromorphic computing frameworks.




%


\bibliographystyle{IEEEtran}
\bibliography{bibliography.bib}

\end{document}